\documentclass[10pt,twocolumn,letterpaper]{article}

\usepackage{iccv}              

\usepackage{wrapfig}
\usepackage{tikz}
\usepackage{multirow}
\usetikzlibrary{positioning}

\definecolor{iccvblue}{rgb}{0.21,0.49,0.74}
\usepackage[pagebackref,breaklinks,colorlinks,allcolors=iccvblue]{hyperref}

\def\paperID{5920} 
\def\confName{ICCV}
\def\confYear{2025}

\def\oursName{EmotiCrafter}
\title{\oursName: Text-to-Emotional-Image Generation based on \\Valence-Arousal Model}

\author {
    Shengqi Dang\textsuperscript{ 1,2}$^{\dagger}$,
    Yi He\textsuperscript{ 1}$^{\dagger}$,
    Long Ling\textsuperscript{ 1},
    Ziqing Qian\textsuperscript{ 1},
    Nanxuan Zhao\textsuperscript{ 3},
    Nan Cao\textsuperscript{ 1,2}$^{*}$ \\\\
    \textsuperscript{1}Tongji University \quad 
    \textsuperscript{2}Shanghai Innovation Institute\quad \textsuperscript{3}Adobe Research
}
\begin{document}
\twocolumn[{%
\renewcommand\twocolumn[1][]{#1}%
\maketitle
\begin{center}
    \centering
    \captionsetup{type=figure}
    \includegraphics[width=\textwidth]{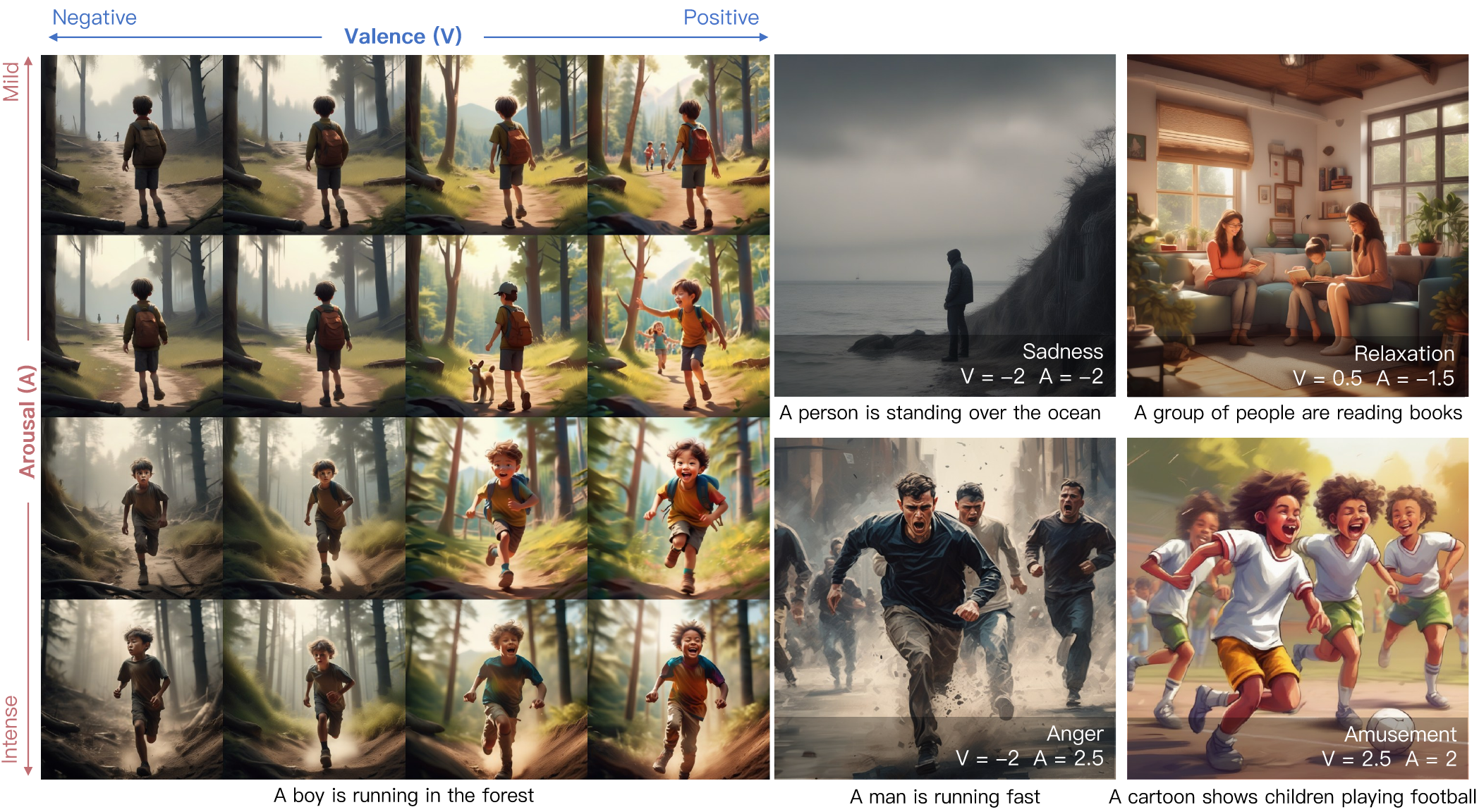}
    \vspace{-0.5cm}
    \caption{Generated images of the\textit{~\oursName}. Given a prompt and Valence-Arousal~(V-A) values, our method can generate an emotional image that reflects the input prompt and aligns with the specified emotion values. By adjusting the V-A values, our method can also generate images that evoke discrete emotions, such as sadness, relaxation, anger, and amusement.}
    
    \label{fig:teaser}
\end{center}%
}]

\maketitle
\renewcommand{\thefootnote}{\fnsymbol{footnote}} 
\footnotetext[0]{$^{\dagger}$Shengqi Dang and Yi He contributed equally to this work.}
\footnotetext[0]{$^*$Nan Cao is the corresponding author.}
\footnotetext[0]{$^\star$Shengqi Dang, Yi He, Long Ling, Ziqing Qian, and Nan Cao are with the Intelligent Big Data Visualization Lab, Tongji University. Shengqi Dang and Nan Cao are also with the Shanghai Innovation Institute. Email: \{dangsq123, heyi\_11\}@tongji.edu.cn, lucyling0224@gmail.com, 2411920@tongji.edu.cn,  nan.cao@gmail.com.}

\footnotetext[0]{$^\star$Nanxuan Zhao is with Adobe Research. Email: nanxuanzhao@g- mail.com.}
\footnotetext[0]{$^\star$Code:~\url{https://github.com/idvxlab/EmotiCrafter}}



\begin{abstract}
Recent research shows that emotions can enhance users' cognition and influence information communication. While research on visual emotion analysis is extensive, limited work has been done on helping users generate emotionally rich image content.
Existing work on emotional image generation relies on discrete emotion categories, making it challenging to capture complex and subtle emotional nuances accurately. Additionally, these methods struggle to control the specific content of generated images based on text prompts.
In this paper, we introduce the task of continuous emotional image content generation (C-EICG) and present \oursName, a general emotional image generation model that generates images based on free text prompts and Valence-Arousal (V-A) values. It leverages a novel emotion-embedding mapping network to fuse V-A values into textual features, enabling the capture of emotions in alignment with intended input prompts. A novel loss function is also proposed to enhance emotion expression. The experimental results show that our method effectively generates images representing specific emotions with the desired content and outperforms existing techniques.

\end{abstract}    

\section{Introduction}
\label{sec:intro}

Emotions are fundamental to human experiences and play a critical role in shaping how people perceive and interact with the world. Research has shown that emotions affect memory~\cite{megalakaki2019effects, lee1999effects, pawlowska2011influence,xie2017negative} and comprehension~\cite{li2020can,vega1996representation,tyng2017influences}, which are crucial for effective communication. As a result, content creators increasingly recognize the importance of incorporating emotions to enhance audience engagement. 

While research on visual emotion analysis is extensive~\cite{yang2017joint,kragel2019emotion,zhao2014exploring}, there is limited work on generating emotionally rich image content. Some early studies explored emotional content generation techniques within specific domains such as facial expressions~\cite{azari2024emostyle, xu2023high} or landscapes~\cite{park2020emotional}. EmoGen~\cite{yang2024emogen} generates images based on a given emotion tag (e.g., happy or sad), which pioneered the domain-free emotional image content generation (EICG) task. However, it has two critical limitations: (1) the image is generated from an emotion tag instead of a text prompt, making the generated content difficult to control; (2) while the discrete emotion tags used in EmoGen are easy to understand, psychologists have not achieved a consensus on the emotion categories~\cite{9472932}. The limited scope of discrete emotion tags falls short of capturing nuanced emotions.

\begin{figure}[t]
    \centering
    \includegraphics[width=\linewidth]{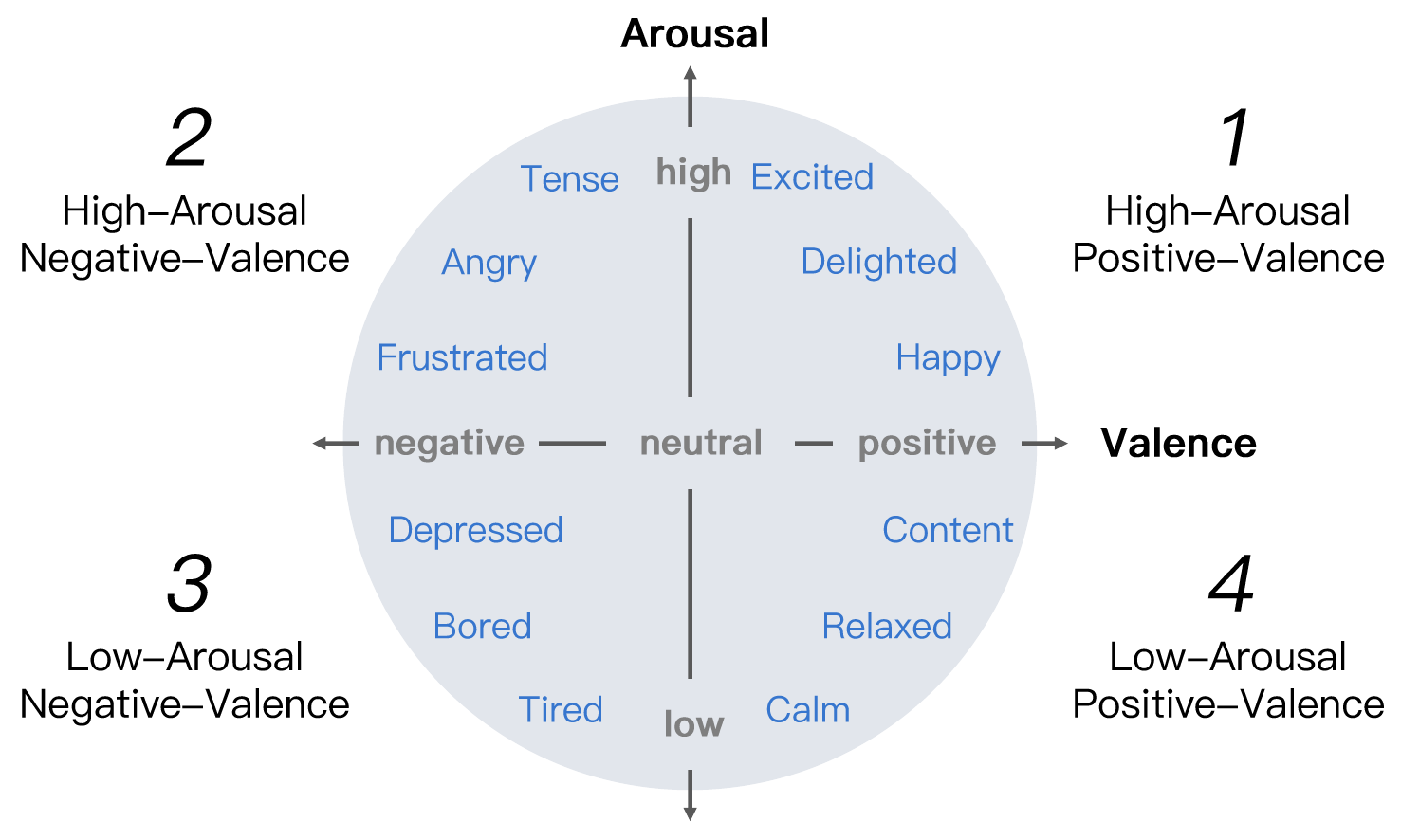}
    \vspace{-5mm}
    \caption{Valence-Arousal model.}  
    \label{fig:vamodel}
\end{figure}

To address the above issues, we propose the continuous emotional image content generation (C-EICG) task and present \textit{\oursName}, the first C-EIGC model that generates emotional images using free-text prompts and continuous Valence-Arousal (V-A) values defined in the V-A model (a well-known psychological continuous emotion model) ~\cite{russell1980circumplex}. V-A model represents emotions in a two-dimensional Cartesian space (Figure~\ref{fig:vamodel}), where Valence quantifies pleasantness (negative to positive) and Arousal measures intensity (calm to excited).  
{We utilize a [-3, 3] range for Valence and Arousal~\cite{khota2022modelling,mertens2024findingemo}. The continuous V-A space enables smooth transitions and nuanced emotional shifts beyond the capability of discrete labels~(e.g., shifts from “bored” to “tired” of a character’s state in a video). Specific emotional values in this space have been investigated in prior work~\cite{evidence}, and such fine-grained modeling is particularly beneficial for human–computer interaction~\cite{azari2024emostyle}. Leveraging this model, \textit{\oursName} captures subtle affective variations via precise (V, A) positioning. The contributions of this paper are as follows:
}
\begin{itemize}
    \item We propose a novel task, continuous emotional image content generation (C-EICG), and develop the first dedicated model for this purpose. Our model introduces an emotion-embedding network that integrates continuous Valence-Arousal (V-A) values with text prompts. These fused features are then injected into Stable Diffusion XL~\cite{sdxl} using cross-attention mechanisms, enabling precise control over both content and emotional expressions in the generated images.
    \item We propose a novel loss function that enhances the emotional resonance of generated images. By amplifying the difference between neutral and emotional text features, our approach enables the model to capture more distinct emotional variations. Additionally, the loss function incorporates the V-A distribution to address data imbalance, further refining the model's ability to generate images with rich, accurate emotional expressions. 
    \item  We constructed an emotional prompts dataset to train the emotion-embedding network, where each sample consists of a neutral prompt and an emotional prompt that share the same core meaning but express a specific emotion corresponding to a given pair of V-A values.
\end{itemize}

\section{Related work}
In this section, we present a review of the related work, specifically focusing on visual emotion analysis, image emotion transfer, and conditional image generation.
\label{sec:related_work}
\subsection{Visual Emotion Analysis}
Visual emotion analysis refers to the computational recognition and interpretation of human emotion~\cite{zhang2024refashioning} in visual media, such as images or videos. 
It has been a prominent research area, with most efforts focusing on the classification of discrete emotions~\cite{9246699,10.1007/978-3-031-19836-6_9,wang2022systematic,borth2013large,xu2022mdan}. However, discrete emotion categories limit the ability to capture the nuanced emotions~\cite{5771357}, leading to an increased focus on continuous emotion analysis in images~\cite{10.1145/2733373.2806354}.

Much of the continuous emotion analysis remains centered on facial expression analysis~\cite{toisoul2021estimation,8013713,zhang2014representation,9881519, Kollias_2022_CVPR}. While effective, facial analysis can overlook crucial contextual information that influences emotional interpretation. This drives studies to focus on emotions in objects or individuals within their environment, rather than just faces\cite{mertens2024findingemo,Kosti_2017_CVPR,meng2022multi,kragel2019emotion}.
For example, Kosti \textit{et al.} \cite{Kosti_2017_CVPR} combined person-specific characteristics with the context of the scene to predict continuous emotional dimensions such as arousal, valence, and dominance. Kragel \textit{et al.} \cite{kragel2019emotion} proposed EmoNet to extract visual features such as facial expressions, body posture, and scene elements from images to predict valence and arousal values.
Recently, Mertens \textit{et al.}\cite{mertens2024findingemo} 
tested multiple backbones (such as ResNet, CLIP, DINO, etc.) for prediction of valence and arousal, demonstrating strong performance across various architectures.

Previous studies have achieved high accuracy in predicting continuous emotions in images, demonstrating a correlation between continuous emotions and visual elements. Building on this insight, our work moves beyond prediction to explore how continuous emotions can be actively embedded within generated content. 

\subsection{Image Emotion Transfer}
Image Emotion Transfer (IET) focuses on editing the content of the images to evoke different emotions \cite{LIU201816,zhu2023emotional,peng2015mixed,weng2023affective}. For instance, Peng \textit{et al.} \cite{peng2015mixed} achieved emotion transfer by adjusting color tones and texture-related features. Zhu \textit{et al.} \cite{zhu2023emotional} introduced a method that separates high-level emotion-relevant features (e.g., object shapes and scene layout) from low-level emotion-relevant features (e.g., brightness). By applying GANs, they transferred emotions between images while preserving their original structure.
Building on these methods, Weng \textit{et al.} \cite{weng2023affective} proposed the Affective Image Filter, which uses a multi-modal transformer to process both text and image inputs. 
With the emergence of text-to-image models, IET has expanded into new applications based on instructive commands. For example, EmoEdit \cite{yang2024emoedit} used GPT-4V to build emotion factor trees that map abstract emotions to specific visual elements and employed the InstructPix2Pix model to apply emotion-driven content and color adjustments to images.

Current methods primarily extract specific features, focusing on certain visual elements or emotional cues, which can limit the depth of emotional expression. In contrast, our method broadly learns a variety of emotion-influencing features, and it could accept natural language prompts as input.


\subsection{Conditional Image Generation}
Conditional image generation aims to create images that align with specific input conditions, such as text~\cite{sd,sdxl,dalle2,dalle3}, reference images~\cite{ip-adaptor}, subjects~\cite{texutual-inverison,dreambooth}, and depth maps~\cite{controlnet,mou2023t2i}.
To enhance quality, researchers have developed specialized approaches, such as Diffusion Transformers (DiT)\cite{dit}, which use transformer-based diffusion for denoising, and Visual Autoregressive Modeling (VAR)\cite{var}, which encodes images into discrete tokens for autoregressive prediction across scales. However, these state-of-the-art methods rely on discrete labels, limiting flexibility and control. In contrast, our approach offers greater generalization by leveraging Stable Diffusion XL~\cite{sdxl} to generate emotional images from free-text prompts.

Despite these progresses, incorporating emotion as a condition for image generation remains underexplored.  EmoGen~\cite{yang2024emogen} pioneered the generation of emotional image content (EIGC) by mapping emotional features to semantic features to generate emotional images. However, EmoGen struggles to understand natural language, limiting its capacity to effectively control specific content. Its reliance on discrete emotions also limits its practical applicability. 

Our method bridges this gap by embedding continuous emotion into textual features, enabling image generation models to use continuous emotion for emotion control. Unlike label-based methods, our method supports free-text prompts for flexible content control.

\begin{figure*}[ht]
    \centering
    \includegraphics[width=\linewidth]{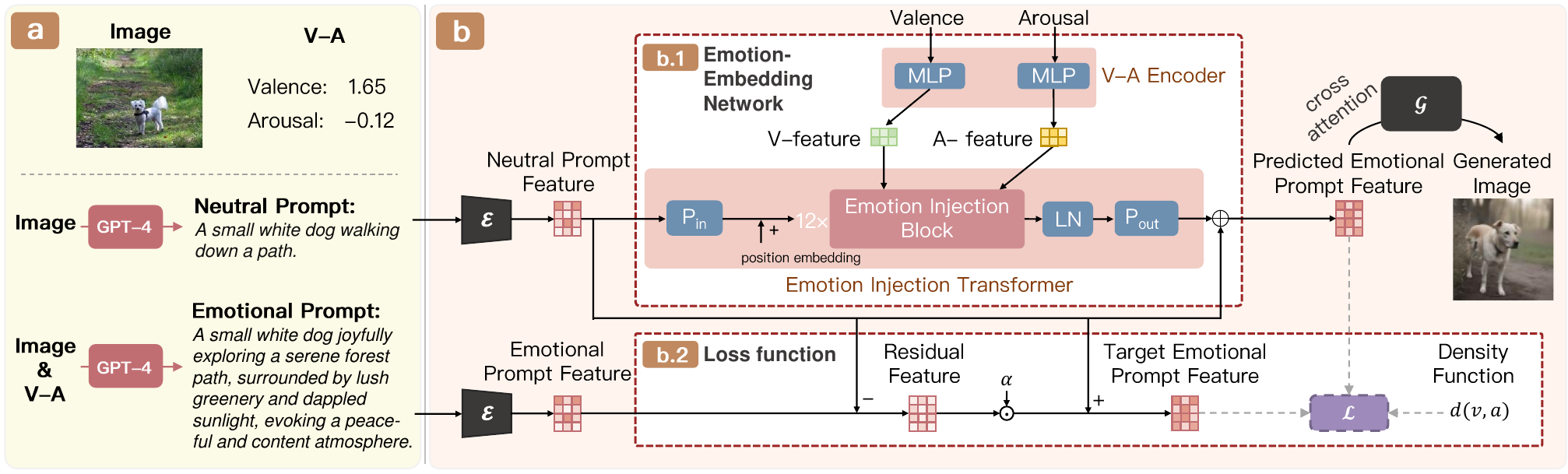}
    \vspace{-5mm}
    \caption{Overview of our method. Specifically, we take the following steps.~(a) We collect an image dataset annotated with V-A values, neutral prompts, and emotional prompts. These prompts are then encoded into features by prompt encoder $\mathcal{E}$.~(b) Next, we design~(b.1) an emotion-embedding network $\mathcal{M}$ to embed V/A values into textual features based on the transformer architecture, and~(b.2) a specialized loss function to enhance the emotional resonance of generated images. The output of the mapping network serves as the condition for the image generation model $\mathcal{G}$ to generate emotional images.}  
    \label{fig:pipeline}
    \vspace{-5mm}
\end{figure*}
\begin{figure}[ht]
    \centering
    \includegraphics[width=\linewidth]{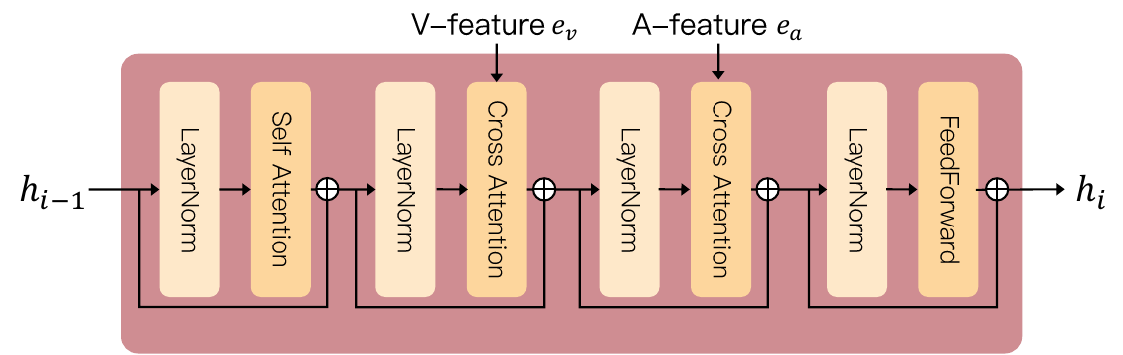}
    \vspace{-5mm}
    \caption{Structure of Emotion Injection Block. It accepts hidden state $h_{i-1}$ as input and produces $h_{i}$ as output. The V-feature $e_v$ and A-feature $e_a$ represent the emotion features, which are injected through the cross-attention module.}  
    \label{fig:eib}
    \vspace{-5mm}
\end{figure}

\section{Method}
\label{sec:method}
In this section, we introduce the technique details of the proposed \textit{\oursName}.

\subsection{Overview}\label{sec:overview}
Our method generates emotional images $I_{emo}$ from two inputs (Figure~\ref{fig:pipeline}): a free-text prompt describing the desired content, and a pair of V-A values $(v, a)$ specifying the emotion. First the prompt encoder $\mathcal{E}$ converts the text prompt into feature $f_n$, which is then processed by an emotion-embedding network $\mathcal{M}$ to produce emotional prompt feature $\hat f_e$ that integrate the V-A values:
\begin{equation}
\begin{aligned}
\hat{f}_{e} &= \mathcal{M}(f_n|(v,a))
\end{aligned}
\end{equation}
Next, this feature is injected into Stable Diffusion XL $\mathcal{G}$ via its cross-attention mechanism to generate the emotional images:
$I_{emo} = \mathcal{G}(\hat{f}_{e})$.
To enhance emotional expressiveness, we introduce a loss function (Fig.\ref{fig:pipeline}(b.2)) that leverages the V-A distribution and emphasizes the differences between neutral and emotional prompt features. This ensures that the generated images accurately convey both the intended emotions and content. Additionally, we construct a dataset (Fig.\ref{fig:pipeline}(a)) that pairs neutral and emotional prompts with the corresponding V-A values.

\subsection{Emotion-Embedding Network}
\label{sec:network}
The emotion-embedding network $\mathcal{M}$ generates the emotional prompt feature by integrating a pair of V-A values with a neutral prompt feature (Fig.\ref{fig:pipeline}(b.1)). First, a V-A encoder converts the V-A values into feature vectors. Then, an emotion injection transformer—modified from GPT-2\cite{radford2019language}—fuses these vectors with the neutral prompt feature, preserving the original textual context while infusing emotional content.

\textbf{V-A Encoder.} 
The V-A Encoder converts a pair of V-A values into feature vectors using two separate multilayer perceptrons (MLPs). One MLP processes the Valence value to produce V-feature $e_v$, and the other processes the Arousal value to generate A-feature $e_a$. These features are then fed into the emotion injection transformer network for further emotion infusion.

\textbf{Emotion Injection Transformer (EIT).}
The Emotion Injection Transformer (EIT) leverages a modified GPT-2 architecture to seamlessly integrate V/A-features into textual features. Its process consists of three stages: input projection, emotion injection, and output projection.

First, we project the input neutral prompt feature $f_n$ into the transformer's feature space:
\begin{equation}
    h_0 = \text{P}_\text{in}(f_n) + \text{PE},
\end{equation}
where $h_0$ represents the initial hidden state; $\text{P}_\text{in}(\cdot)$ is a linear projection layer, and $\text{PE}$ denotes positional embedding~\cite{vaswani2017attention}.

In the next, we inject emotion into $f_n$ via 12 sequential \textbf{Emotion Injection Block}s (\textbf{EIB}s) corresponding to 12 transformer blocks. Each block outputs a hidden state:
\begin{equation}
    h_i = \text{EIB}(h_{i-1}, e_v, e_a), \quad i \in \{1,\dots,12\}
\end{equation}
where $h_i$ is the output of the $i$-th EIB$(\cdot)$. As shown in Figure~\ref{fig:eib}, each EIB enhances the transformer block through a cross-attention mechanism:
\begin{align}
    &h'_i = \text{self-attn}(\text{LN}(h_{i-1})) + h_{i-1} \\
    &h^{(v)}_i = \text{cross-attn}(\text{LN}(h'_i), e_v) + h'_i \\
    &h^{(v,a)}_i = \text{cross-attn}(\text{LN}(h^{(v)}_i), e_a) + h^{(v)}_i \\
    &h_i = \text{fnn}(\text{LN}(h^{(v,a)}_i)) + h^{(v,a)}_i
\end{align}
where $h',h^{(v)},h^{(v,a)}$ are the intermediate hidden variables; $\text{LN}(\cdot)$ is the LayerNorm; $\text{self-attn}(\cdot)$ denotes self-attention, employed to capture context dependencies; $\text{cross-attn}(\cdot)$ is cross-attention for injecting $e_v$ and $e_a$; $\text{fnn}(\cdot)$ is a feed-forward network that adapts the complexity of the emotional embedding process.
We also remove the causal mask typically used for next-token prediction from the original transformer model to fit our task.

Finally, the output of the last (i.e., the 12th) injection block $h_{12}$ is projected back to SDXL's prompt feature space via $\text{P}_\text{out}$ (a linear layer and a LayerNorm layer) to predict the residual between emotional and neutral prompt features, which represents a semantic shift between emotional and neutral prompts:
\begin{equation}
    \hat{f}_r = \text{P}_\text{out}(\text{LN}(h_{12}))
\end{equation}
The final emotional prompt feature is obtained by adding this residual to the original neutral prompt feature:
\begin{equation}
    \hat{f}_e = \hat{f}_r + f_n
\end{equation}

The above emotion embedding network is trained by minimizing the averaged expectation of the difference between the predicted emotional prompt feature $\hat{f}_{e} = \mathcal{M}(f_{n}|(v,a))$ and the scaled target emotional prompt feature $f^{t}_{e}$, using the loss function described in Equation~\ref{equ:loss}.

\begin{equation}\label{equ:loss}
    \mathcal{L} = \frac{1}{n}\mathbb{E}\left(\frac{1}{d(v,a)}\|\hat{f}_{e} - f^{t}_{e}\|^2\right)
\end{equation}
where $n$ is the number of feature elements; $\mathbb{E}(\cdot)$ denotes the expectation; $d(v,a)$ is a density function that describes the distribution of V-A values in the training sample.

To effectively embed emotions and address the challenges posed by the uneven distribution of V-A values in the dataset, this loss function incorporates two key strategies to improve the model's performance:

\textbf{Scaled Residual Learning.} To better capture pronounced emotional changes in generated images, we enlarge the target residuals:
\begin{equation}
    f^{t}_{e} = f_{n} + \alpha 
    \underbrace{(f_{e} - f_{n})}_{\text{residual feature}},
\end{equation}
where $f_{e}$ is the emotional prompt feature, $f_{n}$ is the neutral prompt feature, and $\alpha$ is a scale factor, we set its value to 1.5 based on the ablation study. 

\textbf{V-A Density Weighting.} To mitigate the effects of the imbalanced distribution of the training samples, we weigh the loss inversely proportional to the density of training samples in the V-A space. 
The density is estimated using Kernel Density Estimation (KDE)~\cite{chen2017tutorial} with a Gaussian kernel, denoted as $d(v, a)$:

\begin{equation}
    d(v,a) = \frac{1}{n}\sum_{i=1}^{n}K_H((v,a) - (v_i,a_i)),
\end{equation}
where $K_H$ is a 2D Gaussian kernel with bandwidth $H$; $n$ is the number of training samples; $(v_i,a_i)$ are the V-A values of the $i$-th training sample. The bandwidth $H$ is selected using Silverman's rule of thumb to provide optimal smoothing of the density estimation.

\subsection{Dataset and Training}
\label{sec:dataset}
To train the emotion-embedding network, we constructed a dataset of paired neutral and emotional prompts with corresponding Valence-Arousal (V-A) values (Figure~\ref{fig:pipeline}(a)). These pairs were automatically generated using GPT-4 based on 39,843 images with human-annotated V-A values from publicly available datasets, including OASIS~\cite{OASIS}, EMOTIC~\cite{Kosti_2017_CVPR}, and FindingEmo~\cite{mertens2024findingemo}. Specifically, GPT-4 was used to generate neutral prompts with objective image descriptions and emotional prompts that emphasize affective attributes such as color, lighting, and texture, which influence emotional perception. To ensure data reliability, all LLM-generated prompts underwent crowd-sourced verification, with disagreements resolved through a voting mechanism among annotators.


The proposed emotion embedding network is trained on two NVIDIA A800 GPUs using the aforementioned dataset. We employ the AdamW~\cite{loshchilov2017decoupled} optimizer with a weight decay of 1e-5 and a learning rate of 1e-3. The training process spans 200 epochs with a batch size of 768, completing in approximately 7 to 8 hours.

\begin{figure*}[tbp]
    \centering
    \begin{tikzpicture}
        \node[anchor=north west, inner sep=0] (a) at (0.04\linewidth,0) {
            \includegraphics[width=0.102\linewidth,height=0.102\linewidth]{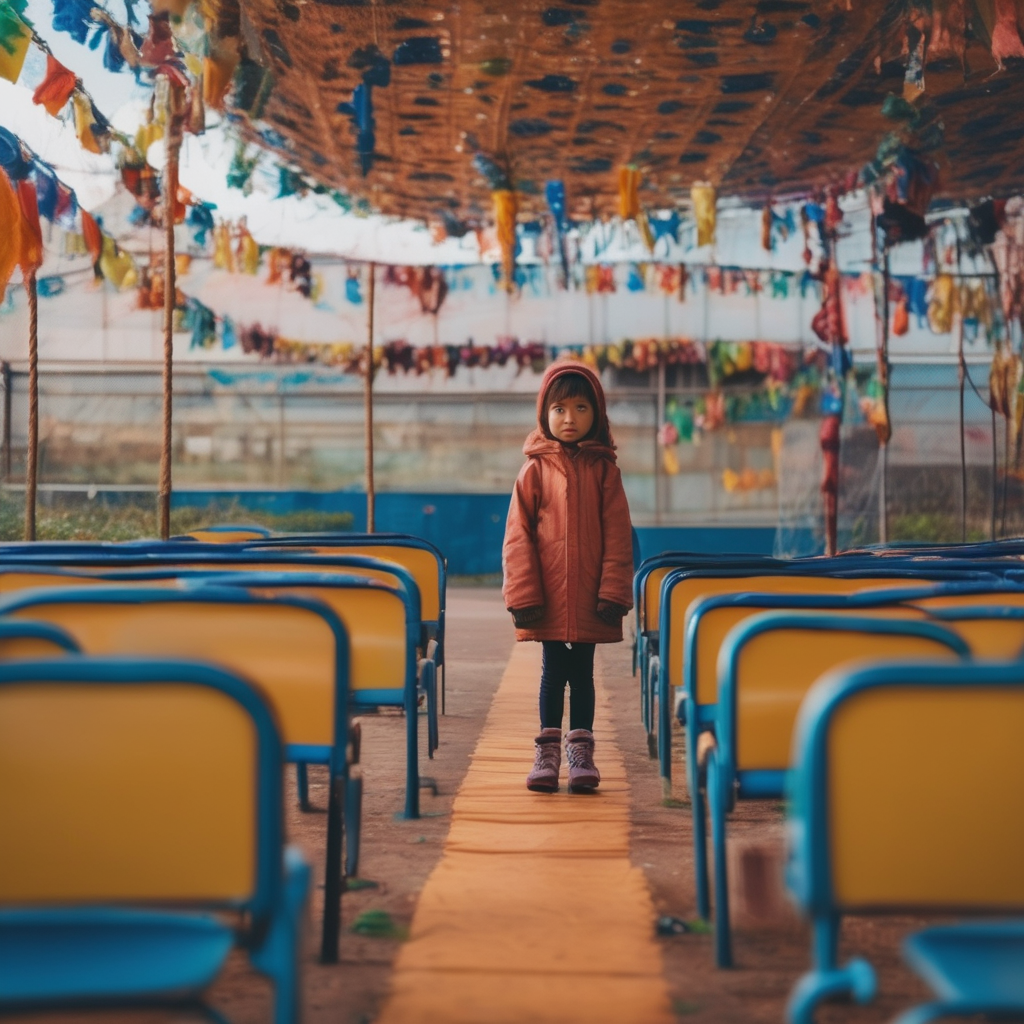}
        };
        \node[anchor=south west] at (0\linewidth,-0.07\linewidth) {(a)};
        \node[anchor=north west, inner sep=0] (b) at (0.184\linewidth,0) {
\includegraphics[width=0.816\linewidth,height=0.102\linewidth]{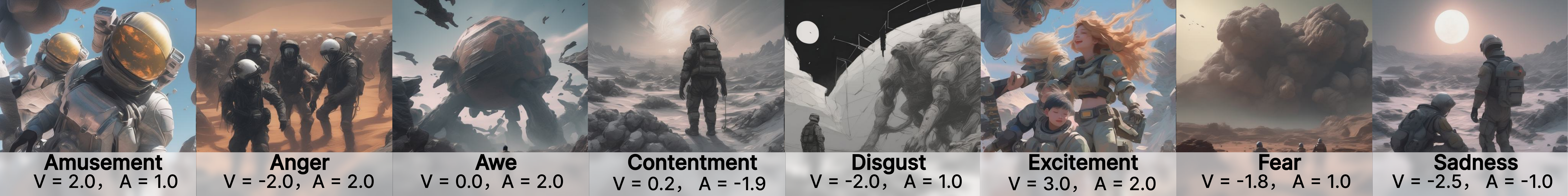}
        };
        \node[anchor=south west] at (0.146\linewidth,-0.07\linewidth) {(b)};
        
        \node[anchor=north west, inner sep=0] (c) at (0.04\linewidth,-0.11\linewidth) {
            \includegraphics[width=0.408\linewidth,height=0.102\linewidth]{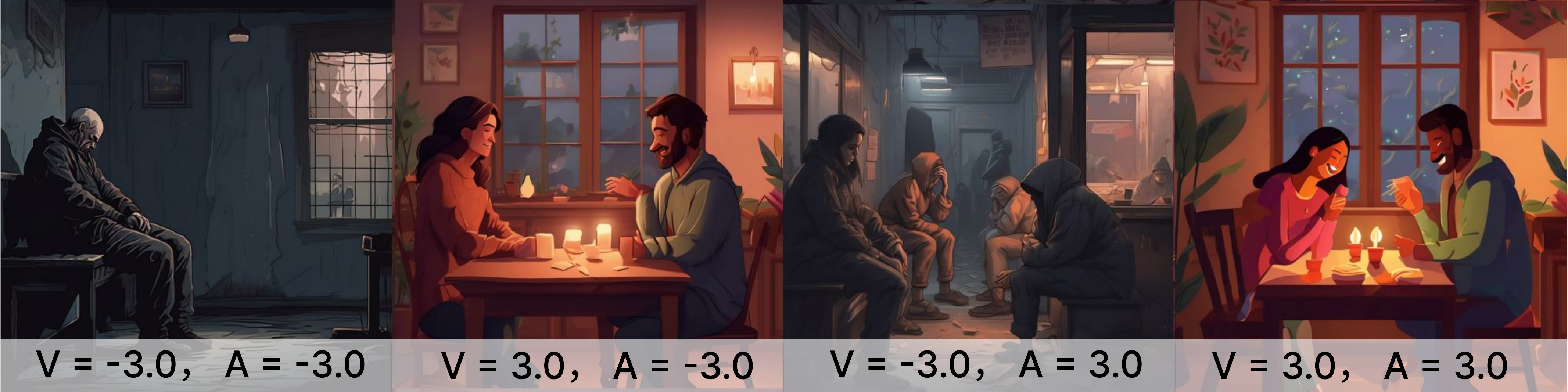}
        };
        \node[anchor=south west] at (0.0\linewidth,-0.18\linewidth) {(c)};
        
        \node[anchor=north west, inner sep=0] (d) at (0.49\linewidth,-0.11\linewidth) {
            \includegraphics[width=0.51\linewidth, height=0.102\linewidth]{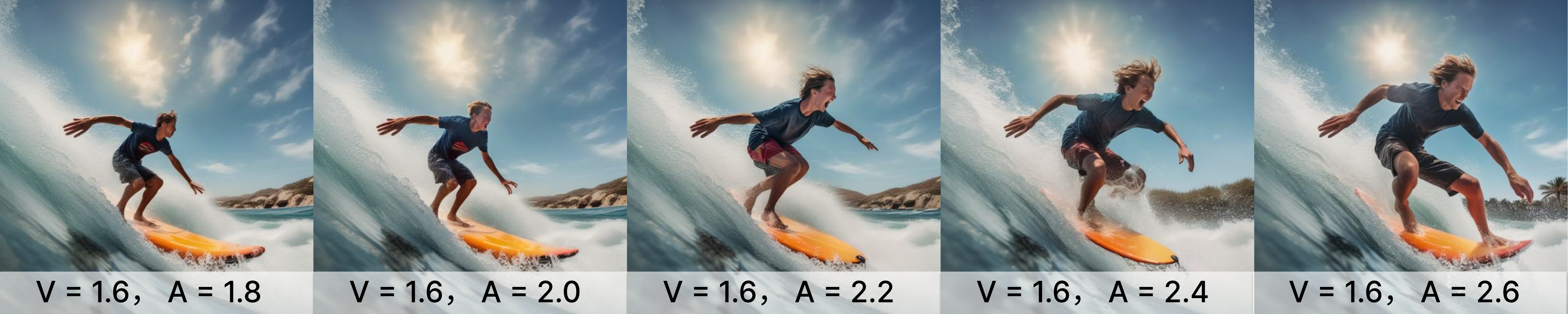}
        };
        \node[anchor=south west] at (0.452\linewidth,-0.18\linewidth) {(d)};
    \end{tikzpicture}
    \vspace{-5mm}
\caption{Results under multiple inputs. (a) Overriding semantic content (`a child in the amusement park') with sad V-A (-2,-2); (b) Discrete emotion mapping in V-A space as emotion input; (c) Empty-prompt generation with pure emotion condition; (d) Fine-grained control of V-A variations with a granularity of 0.2.}

    \label{fig:input-test}
    \vspace{-3mm}
\end{figure*}

\section{Evaluation}

\begin{figure*}[tbp]
    \centering
    \includegraphics[width=0.95\linewidth]{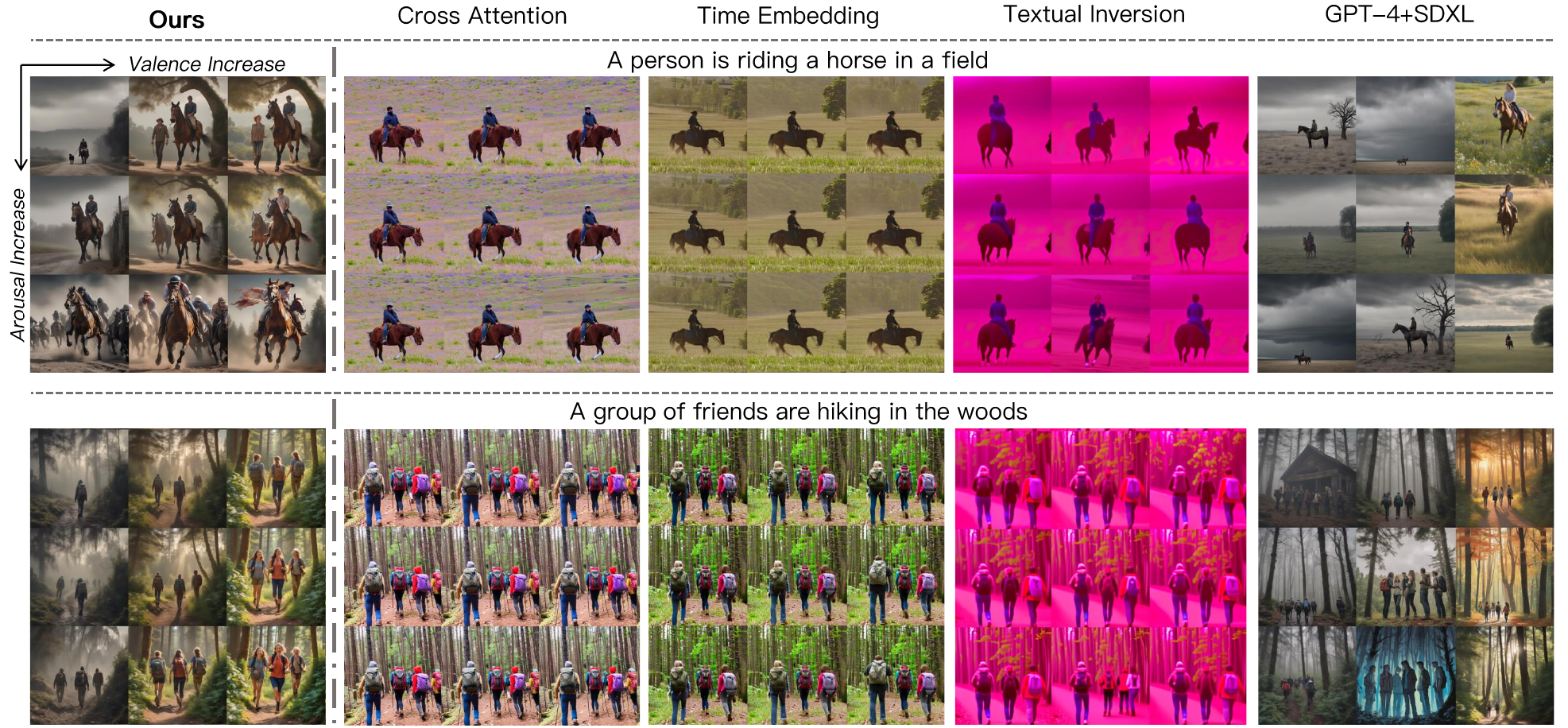}
    \caption{Qualitative comparisons with baselines. These images are generated at varying V-A values, specifically -1.5, 0, and 1.5. Only our approach and the GPT-4+SDXL successfully generate images that clearly reflect emotional variations. Notably, our results show enhanced continuity, indicating superior controllability over continuous V-A values compared to the GPT-4+SDXL.} 
    \label{fig:baseline}
    \vspace{-4mm}
\end{figure*}

\subsection{Generation Results}
Figure~\ref{fig:teaser} shows the proposed technique's ability to achieve continuous and effective control over both emotion and content during image generation. Meanwhile, Figure~\ref{fig:input-test} highlights four key capabilities: (a) emotion-content decoupling, where V-A values override emotional cues in the prompt, allowing typically positive concepts to be rendered with negative emotions; (b) compatibility with discrete emotion categories; (c) content-independent generation, where images generated from empty prompts and specified V-A values maintain emotional consistency without semantic constraints; and (d) fine-grained emotional control, demonstrated through V-A increments of 0.2, showcasing the model's sensitivity to subtle emotional variations.


\begin{figure}[tbp]
    \centering
    \includegraphics[width=\linewidth]{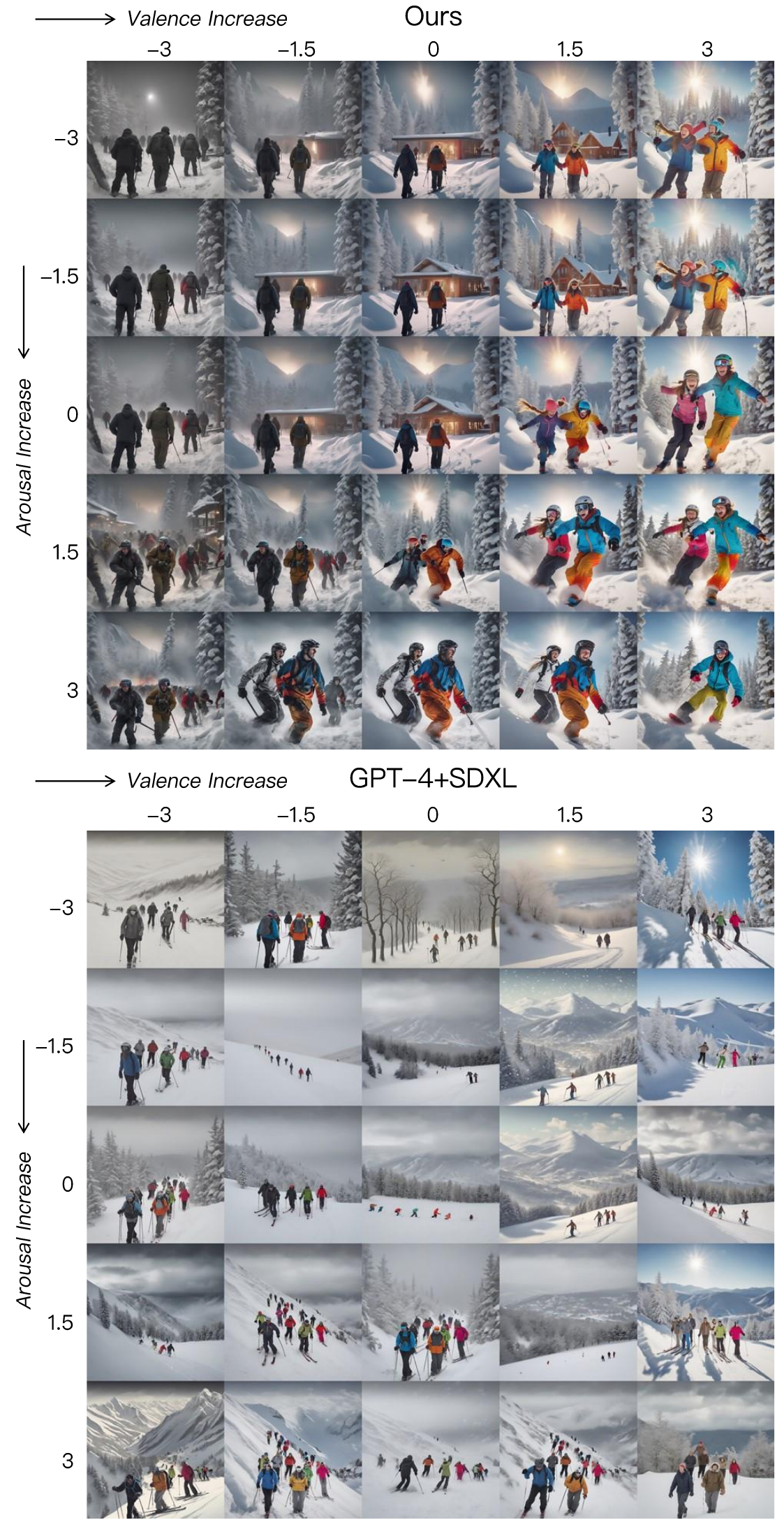}
    \caption{A more comprehensive comparison with the GPT-4+SDXL for the prompt ``A group of people is skiing in a snow hill." Our approach maintains continuity even under extreme V/A conditions. Conversely, GPT-4+SDXL displays noticeable discontinuities (e.g., in the V=3 column).}  
    \label{fig:ours&gpt}
    \vspace{-3mm}
\end{figure}

\subsection{Comparisons}
To estimate the effectiveness of the proposed method, we built four baselines based on existing techniques.

\textbf{Baselines.}
We established baselines for comparison using two strategies: (1) directly injecting emotion features into the image generation modules of SDXL, such as UNet, and (2) modifying the input text prompt using emotion features to influence the generated image's content, similar to the proposed method.

As a result, four different baselines were built: (1) \underline{\textit{Cross Attention}}: inject the emotion features ($e_v, e_a$) into the UNet in SDXL via its cross-attention mechanism based on IP-Adapter~\cite{ip-adaptor}; (2) \underline{\textit{Time Embedding}}: directly add emotion features ($e_v, e_a$) to the time embedding of the UNet in SDXL.
(3) \underline{\textit{Textual Inversion}}: use the text inversion technique~\cite{texutual-inverison} to embed emotion features ($e_v, e_a$) into prompt templates with predefined emotion placeholders. 
(4) \underline{\textit{GPT-4+SDXL (GPT-SD)}}: use GPT-4~\cite{gpt4} to rewrite the input text according to (v,a) values to generate an emotional SDXL prompt for image generation.

\textbf{Qualitative Comparison.} 
We evaluate the generated images based on three criteria: (1) the effectiveness of emotion embedding, (2) image-prompt similarity, and (3) the continuity of emotional variations as V-A values change. {Figure~\ref{fig:baseline} demonstrates that the baseline methods—Cross Attention, Time Embedding, and Textual Inversion—tend to produce nearly identical outputs regardless of emotional variation. This is because the loss terms of these baselines primarily align low-level image features (e.g., SDXL's latent space), which are highly correlated with prompt content but struggle to capture subtle emotional cues when both prompts and V-A values are provided.} Notably, Textual Inversion often produces images with a persistent purple tint. 
Furthermore, all methods faithfully generate content aligned with the given prompts. However, for image continuity, we focus on comparing our method with GPT-4+SDXL, as they are the only two capable of generating distinct emotional variations. Figure~\ref{fig:ours&gpt} provides a more detailed comparison, showing that our method maintains smooth emotional transitions even under extreme V-A conditions, whereas GPT-4+SDXL introduces noticeable discontinuities (e.g., in the V=3 column).

\textbf{\bf Quantitative  Comparison.}
We compare our method against several baselines using the following metrics: (1) \underline{\textit{V/A-Error}} evaluates the absolute error between the predicted V/A of the generated images and the input V/A. 
(2) \underline{\textit{CLIPScore}}~\cite{hessel2021clipscore} assesses the similarity between the input text and the generated images. 
(3) \underline{\textit{CLIP-IQA}}~\cite{clipiqa} leverages a pre-trained CLIP model to evaluate image quality without requiring reference images.
(4) \underline{\textit{LPIPS-Continuous}} utilizes the Learned Perceptual Image Patch Similarity~\cite{zhang2018unreasonable} to measure the continuity of the change in image as V/A changes.

\begin{table}[tbp]
\small
\centering
\resizebox{\linewidth}{!}{%
\begin{tabular}{ccccc}
\hline
 & A-Error $\downarrow$ & V-Error $\downarrow$ & CLIPScore $\uparrow$ & CLIP-IQA $\uparrow$ \\ \hline
Cross Attention & 1.923$\pm$1.153 & 2.080$\pm$1.438 & 26.266$\pm$2.381 & \textbf{0.949$\pm$0.046} \\ \hline
Time Embedding & 1.941$\pm$1.168 & 2.031$\pm$1.348 & \textbf{26.566$\pm$2.125} & 0.786$\pm$0.164 \\ \hline
Textual Inversion & 1.958$\pm$1.188 & 1.923$\pm$1.170 & 22.346$\pm$3.594 & 0.370$\pm$0.111 \\ \hline
GPT-4+SDXL & 1.860$\pm$1.090 & 1.517$\pm$1.060 & 25.907$\pm$1.949 & 0.906$\pm$0.066 \\ \hline
\textbf{Ours} & \textbf{1.828$\pm$1.085} & \textbf{1.510$\pm$1.074} & 23.067$\pm$2.655 & 0.881$\pm$0.099 \\ \hline
\end{tabular}%
}
\vspace{-3mm}
\caption{
Comparison on emotion accuracy, prompt fidelity, and image quality across different baselines, evaluated on 3,300 images per method (132 prompts × 5 V values × 5 A values). Our method achieves the highest performance in emotion accuracy while maintaining comparable results in prompt fidelity and image quality. The slight decrease in prompt fidelity is expected, as modifying emotional content affects semantic alignment.}
\label{tab:quantitative-results}
\vspace{-3mm}
\end{table}

As shown in Table~\ref{tab:quantitative-results}, our method achieves the lowest (best) \underline{\textit{V/A-Error}} on average. While Cross Attention and Time Embedding achieve the highest \underline{\textit{CLIP-IQA}} and \underline{\textit{CLIPScore}}, respectively, these methods fail to generate emotionally expressive images (Figure~\ref{fig:baseline}).

Our method exhibits a slight decrease in \underline{\textit{CLIPScore}} compared to the baselines, which we attribute to the inherent trade-off between emotional modulation and strict semantic alignment. However, the high \underline{\textit{CLIP-IQA}} score indicates that our method produces high-quality images. Additionally, as shown in Table~\ref{tab:ours&gpt}, our approach demonstrates superior continuity compared to GPT-4 + SDXL.

\begin{table}[tbp]
\small
\centering
\begin{tabular}{ccc}
\hline
& \textbf{Ours}  & GPT-4+SDXL \\ \hline
LPIPS-Continuous$\downarrow$&\textbf{0.220$\pm$0.064}&0.361$\pm$0.059
\\ \hline
\end{tabular}
\vspace{-2mm}
\caption{Continuity comparison between our method and baseline.}
\vspace{-3mm}
\label{tab:ours&gpt}
\end{table}

\subsection{User Study}
We conducted two user studies with 20 college students to evaluate the effectiveness of our method by comparing it to \underline{\textit{GPT-4+SDXL}} (the baseline).
 









\begin{table}[tbp]
\centering
\footnotesize 
\vspace{-1mm}
\begin{tabular}{clcc}
\toprule

 &  & \textbf{Ours} & \textbf{GPT-4+SDXL} \\
\midrule
\multirow{4}{*}{Study I} & A-Ranking Consistency $\uparrow$  & \textbf{0.759±0.273} & 0.165±0.379 \\
                        & V-Ranking Consistency $\uparrow$  & \textbf{0.887±0.245} & 0.584±0.259 \\
                        & A-Error $\downarrow$              & \textbf{1.327±1.120} & 2.029±1.446 \\
                        & V-Error $\downarrow$              & \textbf{0.692±0.682} & 1.229±1.026 \\
\midrule
\multirow{2}{*}{Study II} & Emotion Consistency $\uparrow$     & \textbf{4.215±0.715} & 3.525±1.065 \\
                        & Emotion Smoothness $\uparrow$      & \textbf{4.240±0.828} & 3.195±1.163 \\
\bottomrule
\end{tabular}
\vspace{-2mm}
\caption{Our method outperformed the baseline.}
\label{tab:user study}
\end{table}
\textbf{Study I.} In the first experiment, we evaluated whether the generated images’ emotions align with human perception. Two image collections were prepared—one for Arousal (A) and one for Valence (V)—each comprising 20 sets (10 from our method and 10 from the baseline) of 5 randomized images with varying A or V values. Participants reordered the images by perceived intensity and estimated each image's V and A values. We then computed Kendall’s $\tau_b$ (with $\tau_b = 1$ indicating perfect alignment) to assess ordering accuracy and calculated the absolute error between the estimated and ground truth values.

\begin{figure}[tbp]
    \centering
    \vspace{-2mm}
    \includegraphics[width=\linewidth]{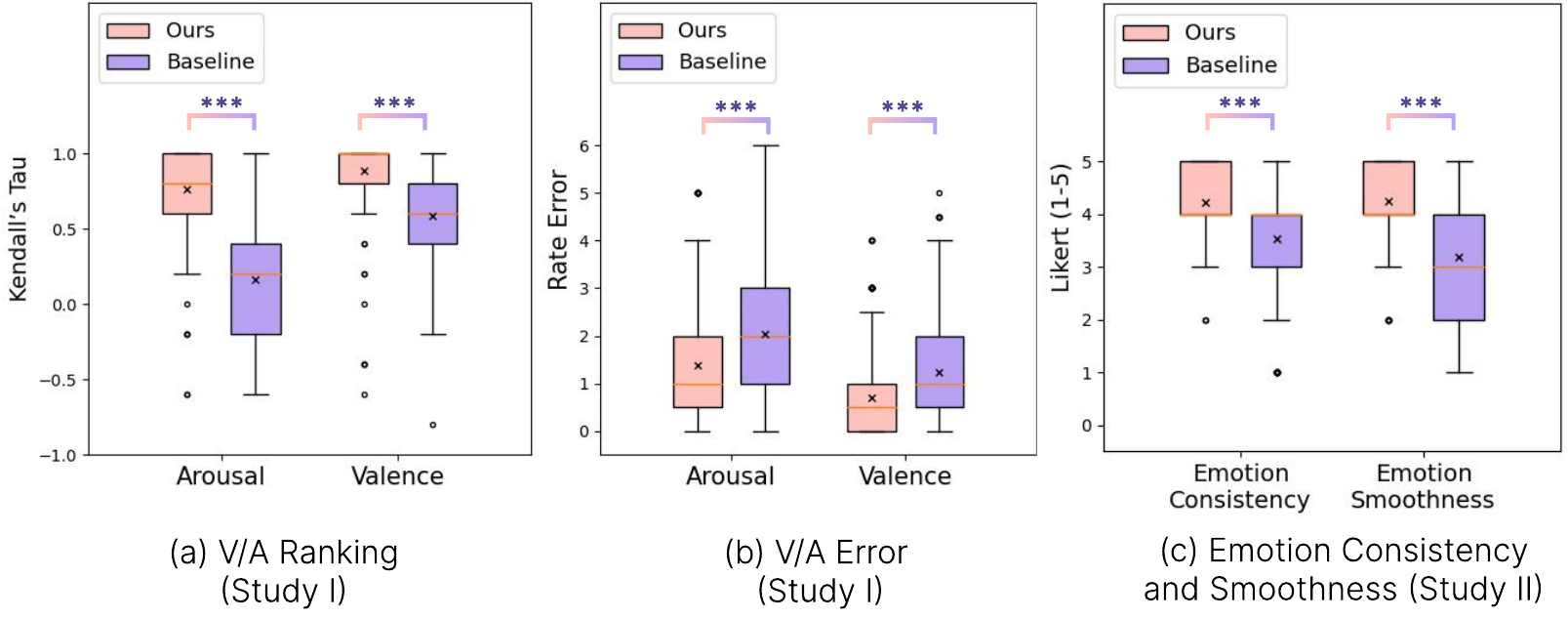}
    \vspace{-7mm}
    \caption{User Study Results ( *p\textless0.05, **p\textless0.01, ***p\textless0.001)}  
    \label{fig:boxplot}
    \vspace{-3mm}
\end{figure}

\begin{figure*}[tbp]
    \centering
    \includegraphics[width=\linewidth]{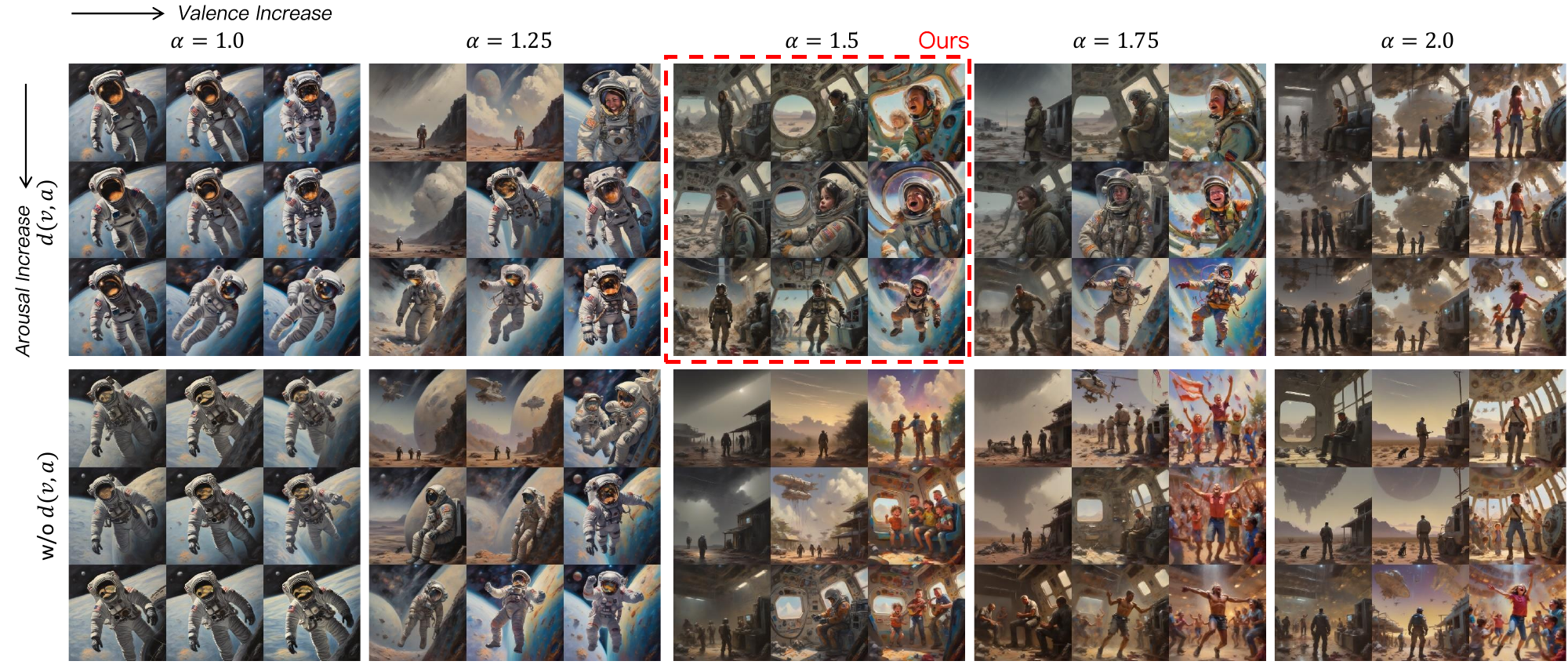}
    \vspace{-3mm}
    \caption{Abalation Study. Images are generated from the prompt ``An oil painting shows an astronaut." As $\alpha$ increases, image-prompt similarity decreases, while emotional variations increase. The usage of $d(a,v)$ enhances the accuracy of emotional changes.}  
    \label{fig:ablation}
    \vspace{-5mm}
\end{figure*}

\begin{figure}[tbp]
    \centering
    \includegraphics[width=\linewidth]{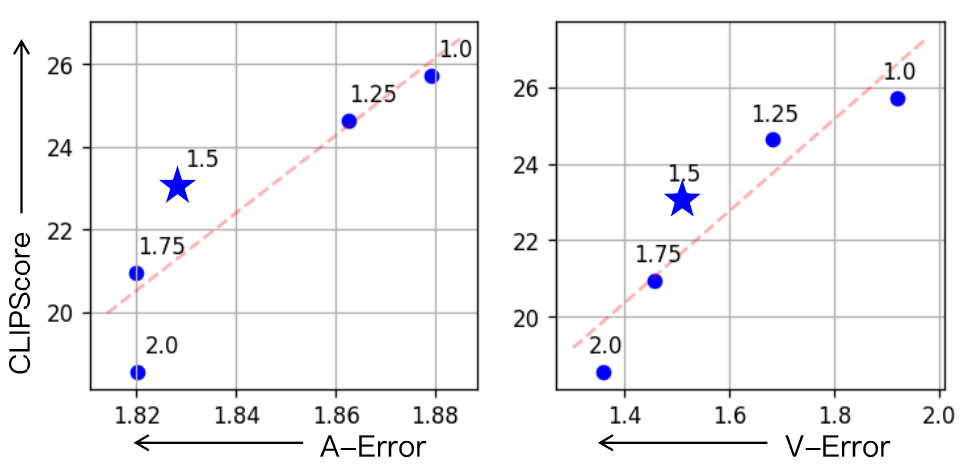}
    \vspace{-6.5mm}
    \caption{The effectiveness of the scaling factor $\alpha$. The method with $\alpha=1.5$ (\textcolor{blue}{$\bigstar$}) surpasses the performance indicated by the regression line (\textcolor{pink}{- - -}).
    } 
    \vspace{-2mm}
    \label{fig:ablation-alpha}
\end{figure}

\textbf{Study II.} In the second experiment, we assessed whether the generated images could effectively reflect continuous emotional changes (i.e., V-A values). We generated 20 image sets (10 per method), each containing 25 images with V-A values varying gradually from -3 to +3 ( Figure~\ref{fig:ours&gpt}). Participants rated each set on a 5-point Likert scale regarding (1) the alignment between V-A changes and image content, and (2) the smoothness of the content transition.

\textbf{Analysis \& Results.} We conducted a Shapiro-Wilk test~\cite{shaphiro1965analysis} to assess normality and applied the Wilcoxon Signed Rank test~\cite{wilcoxon1992individual} to evaluate statistical significance. Our results indicate that our method outperformed the baseline across all metrics (Table~\ref{tab:user study}), with statistically significant improvements in V/A Ranking, V/A Error, Consistency, and Smoothness (Figure~\ref{fig:boxplot}).

\subsection{Ablation Study}
We performed ablation experiments to assess the contribution and effectiveness of the proposed loss function.

\textbf{Effectiveness of the Scaling Factor $\alpha$.} We evaluated how varying $\alpha$ affects CLIPScore and V/A-Error (Figure~\ref{fig:ablation-alpha}). As $\alpha$ increases, CLIPScore decreases (indicating reduced semantic alignment), while V/A-Error also decreases (indicating improved emotional accuracy). This trend is further illustrated by the examples in Figure~\ref{fig:ablation}. Based on these findings, we set $\alpha = 1.5$ as an optimal trade-off, though users can adjust $\alpha$ to suit their specific needs.

\begin{table}[tbp]
\centering
\small
\begin{tabular}{cccc}

\hline
& A-Error $\downarrow$ & V-Error $\downarrow$ & CLIPScore $\uparrow$\\ \hline

Ours    &      \textbf{1.828±1.085}
&\textbf{1.510±1.074} &\textbf{23.067$\pm$2.655
}
   \\ \hline
w/o $d(v,a)$    &    1.829±1.083	

     &    1.546±1.082 &21.977$\pm$0.066

    \\ \hline
\end{tabular}
\vspace{-0.3cm}
\caption{The effectiveness of $d(v,a)$.}
\label{tab:w/o-d(v,a)}
\vspace{-0.2cm}
\end{table}

\textbf{Effectiveness of the Density Weighting $d(v,a)$.} We compare our full method with a variant that omits d(v,a) from the loss function. As shown in Table~\ref{tab:w/o-d(v,a)}, including d(v,a) leads to improvements in both CLIPScore and V/A-Error. This positive effect is further illustrated in Figure~\ref{fig:ablation}.

\section{Conclusion and Limitations}
In this paper, we introduce continuous emotional image content generation (C-EICG) and present \textit{\oursName}, a novel method that generates emotionally expressive images using continuous Valence-Arousal (V-A) values. Our emotion-embedding network integrates V-A values into textual features, and extensive experiments show that our approach reliably aligns images with both user prompts and specified emotions. We believe this work will advance affective computing and image generation, and we will release our code and data to foster further research.

Although our method achieved promising results, it still has some limitations need to be addressed in the future. First, controlling image generation based on arousal remains more challenging than controlling based on valence. This is consistent with prior research in visual emotion analysis, which has found that arousal is harder to predict due to lower inter-annotator agreement~\cite{mertens2024findingemo}. Second, our approach frequently generates images featuring human activities even when such activities are not mentioned in the prompts. This likely stems from the limited representation of non-human scenes in our training data and could be mitigated by incorporating a more diverse range of non-human scenarios. Third, our method occasionally modifies users' input prompts to better align with the specified emotional prompts, resulting in a slight semantic shift that affects the generated images. We believe this issue could be addressed by adding a semantic preservation term to the loss function.

\section{Acknowledgments}
Nan Cao is the corresponding author. This work was supported by the National Key Research and Development Program of China (2023YFB3107100).

{
    \small
    \bibliographystyle{ieeenat_fullname}
    \bibliography{main}
}

\end{document}